\documentclass[sigconf]{acmart}
\AtBeginDocument{%
  }

\setcopyright{acmlicensed}
\copyrightyear{2025}
\acmYear{2025}
\setcctype{by-sa}
\acmConference[ICAIF '25]{6th ACM International Conference on AI in Finance}{November 15--18, 2025}{Singapore, Singapore}
\acmBooktitle{6th ACM International Conference on AI in Finance (ICAIF '25), November 15--18, 2025, Singapore, Singapore}\acmDOI{10.1145/3768292.3770371}
\acmISBN{979-8-4007-2220-2/2025/11}

\begin{document}

\title{Optimizing Large Language Models for ESG Activity Detection in Financial Texts}


\author{Mattia Birti}
\email{mattia.birti@unimib.it}
\orcid{0009-0005-6087-4492}
\affiliation{
\institution{Department of Informatics, Systems and Communication, University of Milano-Bicocca}
  \city{Milan}
  \country{Italy}
}

\author{Andrea Maurino}
\email{andrea.maurino@unimib.it}
\orcid{0000-0001-9803-3668}
\affiliation{
\institution{Department of Informatics, Systems and Communication, University of Milano-Bicocca}
  \city{Milan}
  \country{Italy}
}

\author{Francesco Osborne}
\email{francesco.osborne@unimib.it}
\orcid{0000-0001-6557-3131}
\affiliation{
\institution{University of Milano-Bicocca}
  \city{Milan}
  \country{Italy}
}
\affiliation{
  \institution{The Open University}
  \city{Milton Keynes}
  \country{United Kingdom}
}

\renewcommand{\shortauthors}{M. Birti et al.}

\begin{abstract}
The integration of Environmental, Social, and Governance (ESG) factors into corporate decision-making is a fundamental aspect of sustainable finance. However, ensuring that business practices align with evolving regulatory frameworks remains a persistent challenge. AI-driven solutions for automatically assessing the alignment of sustainability reports and non-financial disclosures with specific ESG activities could greatly support this process. Yet, this task remains complex due to the limitations of general-purpose Large Language Models (LLMs) in domain-specific contexts and the scarcity of structured, high-quality datasets.
In this paper, we investigate the ability of current-generation LLMs to identify text related to environmental activities. Furthermore, we demonstrate that their performance can be significantly enhanced through fine-tuning on a combination of original and synthetically generated data. To this end, we introduce ESG-Activities, a benchmark dataset containing 1,325 labeled text segments classified according to the EU ESG taxonomy. Our experimental results show that fine-tuning on ESG-Activities significantly enhances classification accuracy, with open models such as Llama\_7B and Gemma\_7B outperforming large proprietary solutions in specific configurations. 
These findings have important implications for financial analysts, policymakers, and AI researchers seeking to enhance ESG transparency and compliance through  natural language processing techniques.
\end{abstract}




\maketitle

\section{Introduction}
In recent years, driven by the widespread adoption of the Sustainable Development Goals (SDGs), the European Union has introduced principles and regulations aimed at helping organizations integrate environmental, social, and governance (ESG) factors into their operations and strategic decision-making. These initiatives encourage businesses and investors to assess and improve their environmental impact, fostering a more sustainable approach to economic activity~\cite{kostetckaia2022sustainable}. 
As part of this effort, the ESG taxonomy\footnote{\url{https://finance.ec.europa.eu/sustainable-finance/tools-and-standards/eu-taxonomy-sustainable-activities_en}} has been established, offering a standardized framework to support sustainable practices across industries~\cite{dumrose2022disaggregating}. 
This resource enables companies to evaluate their activities in alignment with its criteria and report their performance in non-financial disclosures and sustainability reports. 

However, aligning business practices with regulatory frameworks continues to be a challenge. Financial investors and other stakeholders often struggle to determine which aspects of corporate social responsibility correspond to specific activities regulated in the ESG taxonomy, making it difficult to accurately interpret and effectively utilize these reports.
Indeed, a company can be associated with a vast amount of text that, in theory, can be mapped to various environmental activities described in the taxonomy. This text may include non-financial disclosures, marketing materials, website descriptions, product and service descriptions, corporate sustainability reports, and more.  
The rise of social fintech technologies, which increasingly support intelligent solutions for social impact investing, presents a promising opportunity to leverage advanced computational techniques for analyzing financial texts and extracting sustainability-related insights. For instance, several AI models have been developed to classify financial documents based on relevant SDGs~\cite{guariso2023automatic,hajikhani2022mapping}. The rapid advancement of large language models (LLMs) in recent years has further improved automated solutions for the analysis of financial and sustainability-related texts~\cite{llama,gemma,gemini,mistral,openai2023gpt4o}. However, categorizing text based on specific environmental activities, such as those defined in the ESG taxonomy, remains a significant challenge such as promoting low-carbon rail transport, ensuring green port operations with zero direct CO$_2$ emissions. This task is considerably more complex than linking text to broad SDG categories, as it requires a fine-grained understanding of the actions and activities described in a document. 
General-purpose LLMs struggle with this level of specificity. Overcoming these limitations, therefore, requires fine-tuning such models on high-quality, domain-specific datasets -- an effort often hindered by the scarcity of training data in this domain.

In this paper, we explore the ability of current-generation LLMs to identify text related to environmental actions. Furthermore, we demonstrate how their performance can be significantly enhanced by fine-tuning them on a combination of original and synthetically generated data. Specifically, our study focuses on classifying textual segments extracted from Non-Financial Disclosures (NFDs) according to the activities defined in the ESG taxonomy. These activities include measures such as reducing emissions through energy-efficient production processes, transitioning to renewable energy, and implementing water conservation strategies, such as wastewater treatment and reuse in manufacturing plants.

To fine-tune and evaluate a range of models on their task, we introduce \textit{ESG-Activities}, a novel benchmark consisting of 1,325 texts classified according to ESG activities. Since we intended to assess the capability of synthetic data in supporting AI systems in this domain, the training set of ESG-Activities includes a mixture of manually curated data from human experts and synthetic data generated by LLMs. In contrast, the test set consists exclusively of items curated by human experts to ensure a high-quality evaluation.

In the evaluation, we simulate realistic low-resource scenarios and therefore focus primarily on relatively lightweight LLMs with sizes between 2B and 7B, drawn from several families including LLaMA, Gemma, and Mistral. We also compare these solutions with alternative approaches, such as the proprietary GPT4o-mini and the domain-specific, encoder-only ESG-BERT.

The experiments demonstrate that fine-tuning the system with a combination of manually crafted and synthetically generated data leads to significant improvements, not only compared to zero-shot learning but also over models fine-tuned exclusively on the manually annotated dataset. Furthermore, we show that relatively small open-source models, such as Llama 7B, can achieve excellent performance when fine-tuned on ESG-Activities, even surpassing large proprietary models in certain configurations.

In summary, the contributions of this paper are as follows:

\begin{itemize}
\item We analyze the performance of state-of-the-art LLMs on the challenging task of associating textual descriptions with specific environmental activities. 
\item We introduce ESG-Activities\footnote{ESG-Activities -  \url{https://github.com/Mattia-Brt/Fine_tuning_LLM/tree/main/data} }, a novel benchmark that combines both original and synthetic data.
\item We demonstrate that augmenting a high-quality, manually curated dataset for fine-tuning with synthetic data enhances the performance of the resulting models on this task.
\item We provide the complete codebase for our analysis to ensure reproducibility\footnote{Codebase - \url{https://github.com/Mattia-Brt/Fine_tuning_LLM/tree/main}}.
\end{itemize}
\hspace{0.8cm}

The remainder of this paper is structured as follows. Section \ref{sec2:taskAndUse} discusses the use case and provides a formal definition of the task. Section \ref{sec3:bechamrk} introduces the new benchmark. Section \ref{sec4:methodology} describes the methodology adopted in our analysis. Section \ref{sec5:eval} presents the evaluation and discusses key insights. Section \ref{sec6:sota} reviews related work. Finally, Section \ref{sec7:conclusion} concludes the paper and outlines future research directions.

\section{Task definition and use case}
\label{sec2:taskAndUse}
\subsection{The role of ESG taxonomy for  Social Fintech}

In the European landscape, the analysis of social and environmental impact investments and the assessment of the sustainability of companies, including crowdfunding platforms or fintech, are becoming increasingly crucial. In 2021, the European Commission defined a set of ambitious policies related to the so-called ``European Green Deal'' with the overarching goal of making Europe climate-neutral by 2050. Among other initiatives, the European Green Deal has established a taxonomy that defines the activities companies must undertake to enhance their contributions to ESG aspects. This taxonomy is becoming a crucial benchmark for European investors, who utilize it to assess the sustainability of their investment decisions.  Investors require a thorough understanding of the extent to which companies comply with the European ESG taxonomy. This is essential for making informed investment decisions that align with their values and contribute to sustainable development. However, the complexity of the ESG taxonomy and the vast amount of financial documentation can make it challenging for investors to independently assess a company's compliance. As a result, there is an increasing need for tools that can streamline this process by automatically identifying and annotating relevant sections within financial documents. By enabling investors to identify companies genuinely committed to sustainability, these technologies help direct capital toward initiatives aligned with SDGs, such as those outlined in the European Green Deal. 

Automating ESG analysis with AI enhances transparency and promotes responsible investment opportunities, ultimately fostering a more sustainable and equitable financial system.

\subsection{Task definition}
In this paper, we describe the process of developing a language model tailored to verify if a text portion from NFD relates to a particular item in the ESG taxonomy. The task is framed as a binary classification problem, where the goal is to train the model to answer the question: \textit{Does c pertain to i?} Specifically, for a text segment \( c \) and an item \( i \) -- in this case, the description of an ESG activity --  the model implements the function \( f(c, i) \)  defined as:
$$
f(c, i) =
\begin{cases} 
1 & \text{ if c pertains to  i}, \\
0 & \text{otherwise}.
\end{cases}
$$

This fine-tuning process is performed using a dataset of triples \((c, i, \text{class})\), where \( c \) is a text chunk, \( i \) is an activity description in the ESG taxonomy, and \(\text{class} \in \{0, 1\}\) indicates whether \( c \) is related to \( i \). The goal is to minimize the binary cross-entropy loss:

\[
\mathcal{L} = -\frac{1}{N} \sum_{j=1}^N \left[ y_j \log \hat{y}_j + (1 - y_j) \log (1 - \hat{y}_j) \right],
\]

where \( N \) is the number of samples in the dataset,
\( y_j \) is the true label (\( \text{class} \)) for the \( j \)-th sample, and
\( \hat{y}_j = f(c_j, i_j) \) is the predicted probability that \( c_j \) pertains to \( i_j \).

The fine-tuning process was performed using the Low-Rank Adaptation (LoRA) technique~\cite{heidloff2023efficient}, which updates only a low-rank subspace of the model's pre-trained weight matrices, reducing the number of trainable parameters while maintaining efficiency.

\section{The ESG-Activities Benchmark}
\label{sec3:bechamrk}
To the best of our knowledge, no existing dataset has been explicitly designed to enhance an LLM's ability to classify text based on precise and granular environmental activities, such as those defined in the ESG taxonomy.
To address this gap, we introduce \textit{ESG-Activities}, a novel benchmark dataset.
ESG-Activities links textual segments from NFDs to specific activities defined in the ESG taxonomy. The dataset was constructed following the process outlined below.

First, we selected the NFDs of four major companies within the transportation industry. 
The selected companies were \textit{Ferrovie dello Stato}, the Italian State Railways; \textit{Autostrade per l’Italia}, Italy’s largest company for motorway management and maintenance; \textit{Maersk}, a well-known shipping and logistics company; and \textit{Mundys}, a multinational company specializing in motorway and airport infrastructure.
Next, we selected 12 activities related to the transport industry from the ESG taxonomy. These activities represent various actions that companies can take to promote sustainability in the transportation sector, such as developing zero-emission vessel infrastructure with electric charging and hydrogen refuelling, implementing smart mobility systems to enhance traffic efficiency, promoting low-carbon rail transport with CO$_2$-free or bimodal trains, expanding infrastructure for walking, cycling, and personal electric mobility, and ensuring green port operations with zero direct CO$_2$ emissions.

Since the original descriptions were very long, we generated shorter versions using GPT-4 as a rephraser. We then indexed the NFDs in a vector database and employed a Retrieval-Augmented Generation (RAG)~\cite{lewis2020retrieval} pipeline to retrieve the most relevant text chunks for each activity description. This process yielded 265 candidate mappings between text segments and ESG actions.

The candidate mappings were evaluated by three domain experts -- professors and postdoctoral researchers with expertise in the transport industry. Each annotator independently assessed whether to confirm or reject each candidate mapping. A mapping was deemed valid if it received at least two positive votes out of three, following a majority rule.
The resulting dataset consisted of 265 entries, each containing a textual description, an activity description, and a binary flag indicating a match. Among these, 78 text-activity pairs were classified as true matches. The dataset was then split into a training set of 212 instances and a test set of 53 instances.

Since we aimed to verify the effectiveness of synthetic data in this domain, we then expanded the training set with artificial data.
Specifically, we used ChatGPT-4o to generate five alternative formulations for each of the original 212 sentences in the training set, ensuring they conveyed the same meaning while using different wording. This process resulted in 1,060 additional sentences, which were labelled as 'synthetic data' and incorporated into the training set. This setup allows users of the benchmark to fine-tune their models using only the original data (212 items) or a combination of original and synthetic data (1272 items).

\section{Methodology}
\label{sec4:methodology}
In this section, we first introduce ESGQuest, the prototype we developed for analysing documents and mapping them to ESG activities using an LLM fine-tuned for this task. We then provide an overview of the LLMs evaluated in this study and describe the experimental setup used to assess their performance on the ESG-Activities benchmark.

\subsection{System Architecture}
\label{sec3:architecture}

The study presented in this paper is part of a broader project aimed at developing ESGQuest, a research prototype designed to support the annotation of NFDs with ESG-related activities. This system is built on a RAG pipeline. 
Given a document about a company and a set of relevant NACE codes, ESGQuest aims to identify text segments that align with ESG activities associated with the specified NACE codes. The results are presented as an annotated PDF, which can be further refined by human users.\footnote{A demo of this system is available at \url{https://esgquest.datai.disco.unimib.it}. } 
The ESGQuest pipeline operates as follows.

First, ESGQuest selects ESG activities based on the given NACE codes using a predefined mapping schema. \textit{NACE codes} (Nomenclature statistique des Activités économiques dans la Communauté Européenne) are a European industry standard classification system that groups organizations based on their economic activities.\footnote{ These codes are used in the EU Taxonomy to assess the environmental sustainability of economic activities. For more details, visit \url{https://ec.europa.eu/eurostat/web/nace}} This step is optional but generally beneficial, as many ESG activities described in the taxonomy are relevant only to specific industrial sectors (e.g., transport, energy). Including irrelevant activities in the analysis would not only waste computational resources but also introduce unnecessary complexity and potential confusion. 

Second, the input documents are divided into smaller chunks and stored in Pinecone\footnote{Pinecone - \url{https://www.pinecone.io/}}, a vector database optimized for similarity search. Pinecone facilitates the efficient retrieval of the most semantically relevant segments for a given query by leveraging high-dimensional vector embeddings. It is specifically designed for approximate nearest neighbour search, ensuring fast and scalable retrieval even for large datasets.

Third, for each relevant ESG activity, the system queries the vector database with the activity description to retrieve the most pertinent chunks from the input documents. 

Finally, an LLM optimized for this purpose analyzed each chunk to assess whether it aligned with the description of the action.
If a chunk is assigned to an action, the corresponding text segment in the original PDF is annotated accordingly.

\subsection{Large Language Models}\label{ref:models_list}
In this paper, we examine a range of LLMs, including several open-source models, which offer a cost-effective solution and can be deployed locally to ensure the privacy of sensitive documents, as well as a large proprietary model (GPT-4o Mini). Below, we provide a description of each LLM evaluated in this paper, including their sizes and key characteristics.

\subsubsection{Llama. }
Llama, introduced by Hugo Touvron et al. in 2023~\cite{llama}, is a family of decoder-only language models designed to achieve both high performance and computational efficiency. The Llama models span from 1 to 90 billion parameters, making them suitable for a wide range of natural language processing tasks. They were trained on diverse datasets, including publicly available sources such as CommonCrawl, C4~\cite{t5}, GitHub, Wikipedia, Books3~\cite{books3}, ArXiv, and Stack Exchange.
In this study, we evaluate three models from the Llama family: 
Llama 3B\footnote{\url{https://huggingface.co/openlm-research/open\_Llama\_3b}}, Llama 2 7B\footnote{\url{https://huggingface.co/meta-Llama/Llama-2-7b-hf}}, and Llama 3 8B\footnote{\url{https://huggingface.co/meta-llama/Meta-Llama-3-8B}}.

\subsubsection{Gemma. }
Introduced by Google in 2024, Gemma~\cite{gemma} is a family of open-source language models derived from the Gemini series~\cite{gemini}, designed to achieve an optimal balance between computational efficiency and performance across various natural language processing tasks. 
Gemma employs a decoder-only architecture with advanced features such as multi-query attention to reduce memory overhead \cite{team2024gemma}, Rotary Position Embeddings (RoPE)~\cite{t5} to improve positional representation, GeGLU activations for better convergence during training, and RMSNorm to enhance stability on large-scale datasets. The models were trained on diverse publicly available data, including CommonCrawl, C4~\cite{t5}, GitHub repositories, ArXiv papers, Wikipedia, and Books3~\cite{books3}, ensuring broad linguistic coverage and domain-specific expertise. In this study, we evaluate three models from the Gemma family: 
Gemma\_2B\footnote{\url{https://huggingface.co/google/gemma-2b}}, Gemma\_7B\footnote{\url{https://huggingface.co/google/gemma-7b}}, and RecurrentGemma\_2B\footnote{\url{https://huggingface.co/google/recurrentgemma-2b}}.

\subsubsection{Mistral. }
Mistral, introduced by Albert Q. Jiang et al. in 2023~\cite{mistral}, is a highly efficient decoder-only language model designed for a wide array of natural language processing tasks, excelling in areas such as reasoning, mathematics, code generation, and commonsense reasoning. With 7.3 billion parameters, Mistral was specifically engineered to achieve a balance between computational efficiency and performance. The model incorporates several advanced techniques that make it highly effective for tasks that demand long-context processing and memory optimization. One such technique is the sliding window attention (SWA)~\cite{swa}, which enables the model to process long sequences of input in overlapping chunks, allowing it to handle inputs of arbitrary length with minimal computational overhead. 
In this study, we employ the 7B variant of Mistral\footnote{\url{https://huggingface.co/mistralai/Mistral-7B-v0.1}}, which has been fine-tuned to support a variety of use cases, including scientific research where the accurate interpretation and generation of complex texts are essential. 

\subsubsection{Gpt-4o Mini.}
GPT-4o Mini is a decoder-only language model developed by OpenAI as a faster and more cost-effective alternative to GPT-4o. GPT-4o is a comprehensive multimodal model capable of processing and generating text, images, and audio, designed for complex tasks requiring high computational power and precision. It features a 128K-token context window and supports up to 16K output tokens per request, with a knowledge cutoff in October 2023~\cite{openai2023gpt4o}. 
Despite its smaller size, GPT-4o Mini outperforms previous models, such as GPT-3.5 Turbo, in academic benchmarks, demonstrating superior performance in textual intelligence and multimodal reasoning~\cite{openai2023gpt4omini}. It retains the same context window and output token limits as GPT-4o, ensuring robust performance across various applications. 
Additionally, GPT-4o Mini incorporates an instruction hierarchy method to enhance resistance to jailbreaks and prompt injections, improving security for deployment in diverse settings~\cite{openai2023gpt4osafety}.

\subsubsection{ESG-BERT}
ESG-BERT~\cite{nbroad_ESG-BERT} is a domain-specific variant of BERT, pre-trained on a large corpus of sustainability-related texts. It is designed to capture the terminology and linguistic nuances of the ESG domain, and it achieves strong performance on text classification and sentiment analysis tasks in this context. Specifically, ESG-BERT has been shown to effectively identify key concepts related to sustainability and corporate social responsibility. In this study, we include ESG-BERT as a baseline for comparison with fine-tuned LLMs.

\subsection{Experimental Setup}
In order to study the advantages of fine-tuning LLMs for ESG activity classification, we evaluated the models in two different modes: zero-shot and fine-tuning. The fine-tuning was performed on the two alternative training sets within ESG-activities: the manually curated dataset of 212 instances and a larger dataset of 1,272 instances, which combines original and synthetic data. 
 In both cases, the models were evaluated on the ESG-activities test set using standard metrics for binary classification: precision, recall, and F1-score. We used the weighted-averaged version of these metrics, given the unbalanced nature of the dataset, which contains more instances of class 0 than class 1. 
In the following sections, we provide a detailed discussion of both approaches.

\subsubsection{Zero-shot}
Zero-shot learning is a method in which a LLM is prompted to perform a task based solely on provided instructions, without the need for task-specific examples. This approach leverages the extensive pretraining of LLMs, allowing them to generalize to new, previously unseen tasks. It is particularly valuable because it eliminates the need for additional task-specific training data, making it a cost-effective and scalable solution across various applications.

In this study, we designed and iteratively refined a prompt template for zero-shot classification. The template takes two inputs: 1) the text to be classified and 2) a description of an ESG action. Based on these inputs, the prompt instructs the LLM to return a value of 1 if the text aligns with the ESG action and 0 otherwise.

\subsubsection{Fine-tuning}
Fine-tuning is a supervised learning technique that involves adjusting the weights of a pre-trained LLM using a labelled dataset. This process refines the model’s broad language understanding and knowledge acquired during its initial pre-training phase, enabling it to achieve superior performance on specific tasks or within particular domains.

In this study, we fine-tuned nine models for the proposed task using the training and validation datasets from the ESG-Activities benchmark. Specifically, each model was fine-tuned twice: first using the manually curated training and validation datasets, and then using augmented datasets that included synthetic data. This approach enables a comparative analysis of the two model versions, allowing us to evaluate the impact of synthetic data on models of different sizes.\\
Due to the limited size of our training dataset, we employed a 10-fold cross-validation approach to mitigate the risk of overfitting. This technique systematically partitions the dataset into ten subsets, ensuring that the model is trained and evaluated on different data portions in each iteration. By rotating the validation set across all folds, cross-validation reduces bias and variance, leading to a more reliable assessment of the model’s generalization capability and performance metrics.
Models were then saved after each epoch, and validation was conducted at regular intervals, referred to as \textit{steps}. Specifically, the models were evaluated and saved every 10\% of the total optimization steps. This strategy enables the selection of the best-performing model based on validation metrics and mitigates issues like overfitting or underfitting.\\
The optimization process was handled using the \texttt{AdamW} optimizer \cite{guan2023weightpredictionboostsconvergence}, a variant of the Adam optimizer that incorporates decoupled weight decay regularization. To optimize the fine-tuning process, we employed LoRA techniques from the PEFT library, as previously mentioned. This approach reduces the number of trainable parameters while preserving model performance. For instance, when fine-tuning Llama 3B, which has over 3.3 billion parameters, LoRA required modifying only 0.08\% of the total parameter space. Similarly, for Llama 7B, only 0.12\% of its parameters were updated.

To enhance transparency and reproducibility, we summarize the main experimental settings and hyperparameters used during the fine-tuning process in Table~\ref{table:hyperparameters}. These configurations were applied consistently across all model variants unless explicitly stated otherwise. The settings were chosen based on preliminary experiments and established best practices for low-resource fine-tuning scenarios. 

\begin{table}[h]

\centering
\begin{tabular}{ll}
\hline
\textbf{Parameter} & \textbf{Value} \\
\hline
Gradient accumulation steps & 5 \\
Learning rate & 3e-4 \\
Validation split ratio & 0.15 \\
Cross-validation folds & 10 \\
Evaluation steps & Every 10\% of total steps \\
Optimizer & AdamW \\
Adapter type & LoRA \\
LoRA rank ($r$) & 8 \\
LoRA alpha & 32 \\
LoRA dropout & 0.05 \\
LoRA target modules & \texttt{["query\_key\_value"]} \\
\hline
\end{tabular}
\vspace{2mm}
\caption{Summary of experimental and hyper-parameter settings for fine-tuning.}
\label{table:hyperparameters}
\end{table}

Two models required deviations from the standard fine-tuning procedure described above. GPT‑4o Mini was fine‑tuned using Azure OpenAI\footnote{Azure OpenAI — \url{https://azure.microsoft.com/en-us/products/ai-services/openai-service}}, a cloud‑based service provided by Microsoft that offers access to OpenAI’s language models. We formatted the training dataset in the required JSON structure, uploaded it to Azure, and executed the fine-tuning process with the default hyperparameters.  ESG‑BERT also required an additional step. This model was originally designed as a 26‑class classifier specifically for ESG text classification tasks. Since no publicly available version of the model without the task‑specific 26‑class output layer was available, we adapted the existing checkpoint by removing its final classification layer and replacing it with a binary output layer, making it suitable for the binary classification task defined in this study. 

\section{Evaluation}
\label{sec5:eval}

In this section, we present the results of our experiments on the ESG-Activities benchmark. Specifically, we evaluated the nine models described in Section \ref{sec4:methodology} using three alternative configurations: 1) zero-shot learning, 2) fine-tuning on the original data from ESG-Activities, and 3) fine-tuning on a combination of the original and synthetic data. 
 As previously mentioned, we evaluated the performance of the models using standard metrics for binary classification: precision, recall, and F1-score.

\color{black}

\subsection{Zero-Shot Setting}
Table~\ref{table:ZSL} reports the performance of the models in a zero-shot setting. 
\texttt{GPT4o\_mini} yields the best result (72.7\%), followed by \texttt{Gemma\_7B} (70.5\%), and \texttt{Llama3\_8B} (70.1\%). 
\texttt{Gemma\_2B} and \texttt{Llama\_3B}  performed fairly well, exceeding 68\% F1. 
Conversely, \texttt{ESG-BERT}, \texttt{RecurrentGemma\_2B} and \texttt{Mistral\_7B} showed relatively poor performance in this task. 
Overall, there is considerable room for improvement.

\begin{table}[h]
\centering
\begin{tabular}{lcccc}
\hline
\textbf{Model} & \textbf{Precision} & \textbf{Recall} & \textbf{F1-Score} \\
\hline
\texttt{Llama\_3B} & 0.6565 & \textbf{0.7358} & 0.6816 \\
\texttt{Llama2\_7B} & 0.7145 & 0.5472 & 0.5828 \\
\texttt{Llama3\_8B} & 0.6845 & \textbf{0.7358} & 0.7008 \\
\texttt{Gemma\_2B} & 0.6696 & 0.6792 & 0.6742 \\
\texttt{Gemma\_7B} & \textbf{0.8361} & 0.6792 & 0.7050 \\
\texttt{RecurrentGemma\_2B} & 0.7455 & 0.4717 & 0.4945 \\
\texttt{Mistral\_7B} & 0.5574 & 0.3207 & 0.3396 & \\
\texttt{GPT4o\_mini} & 0.7202 & \textbf{0.7358} & \textbf{0.7270} & \\
\texttt{ESG-BERT} & 0.6994 & 0.5094 & 0.5455 & \\
\color{black}\\
\hline
\end{tabular}
\vspace{2mm}
\caption{Performance of the models in a zero-shot setting.}
\label{table:ZSL} 
\end{table}

\subsection{Fine-tuning}
Table~\ref{table:FT-original} presents the results of the models fine-tuned on the manually curated dataset. The results demonstrate a significant improvement across all models, confirming the efficacy of the ESG-Activities benchmark in optimizing performance for this task.

The top-performing model is again \texttt{GPT4o\_mini}, which attains an F1 score of 80.5\%, followed by \texttt{Llama\_3B}  (78.9\%)  and \texttt{Gemma\_2B} (78.1\%). 
It is interesting that some of the smaller models achieved the best performance after fine-tuning. This is likely due to the fact that the manually curated training set is relatively small. As a result, models tend to overfit quickly, and this effect is even more pronounced in larger models, which overfit earlier and thus experience a drop in performance. Notably, \texttt{ESG-BERT} achieves a high precision (82.7\%) but suffers from a low recall, which leads to a lower overall F1 score. Several LLMs achieve better performance when fine-tuned on ESG-Activities.

\begin{table}[h!]
\centering
\begin{tabular}{lcccc}
\hline
\textbf{Model} & \textbf{Precision} & \textbf{Recall} & \textbf{F1-Score} \\
\hline
\texttt{Llama\_3B} & 0.7866 & 0.7925 & 0.7892 \\
\texttt{Llama2\_7B} & \textbf{0.8361} & 0.6792 & 0.7050 \\
\texttt{Llama3\_8B} & 0.7420 & 0.7170 & 0.7273 \\
\texttt{Gemma\_2B} & 0.7763 & 0.7925 & 0.7813  \\
\texttt{Gemma\_7B} & 0.8146 & 0.6981 & 0.7226 \\
\texttt{RecurrentGemma\_2B} & 0.7763 & 0.4340 & 0.4396 \\
\texttt{Mistral\_7B} & 0.6701 & 0.5472 & 0.5838 & \\
\texttt{GPT4o\_mini} & 0.8015 & \textbf{0.8113} & \textbf{0.8050} & \\ 
\texttt{ESG-BERT} & 0.8272 & 0.6415 & 0.6689 & \color{black}\\
\hline
\end{tabular}
\vspace{2mm}
\caption{Performance of the fine-tuned models (original data).}
\label{table:FT-original}
\end{table}

Table~\ref{table:FT-synthetic} presents the performance of models fine-tuned on the second training set of ESG-activities, which includes a combination of original and synthetic data. This experiment aimed to determine whether incorporating synthetic data during fine-tuning provides a performance advantage.
The results clearly indicate that it does, with most models now achieving F1 scores exceeding 80\%. The best overall result is achieved by \texttt{Llama2\_7B}, with an F1 score of 84.9\%, followed by \texttt{Llama\_3B} at 84.4\% and \texttt{Gemma\_7B} at 82.1\%. \texttt{GPT4o\_mini}, which was fine-tuned using the OpenAI service, did not perform as well in this setting.  As before, \texttt{ESG-BERT} underperformed. 

\begin{table}[h]
\centering
\begin{tabular}{lcccc}
\hline
\textbf{Model} & \textbf{Precision} & \textbf{Recall} & \textbf{F1-Score} \\
\hline
\texttt{Llama\_3B} & 0.8421 & \textbf{0.8491} & 0.8440 \\
\texttt{Llama2\_7B} & \textbf{0.8491} & \textbf{0.8491} & \textbf{0.8491} \\
\texttt{Llama3\_8B} & 0.8290 & 0.7925 & 0.8036 \\
\texttt{Gemma\_2B} & 0.8176 & 0.8302 & 0.8127 \\
\texttt{Gemma\_7B} & 0.8190 & 0.8302 & 0.8211 \\
\texttt{RecurrentGemma\_2B} & 0.8470 & 0.5283 & 0.5452 \\
\texttt{Mistral\_7B} & 0.8032 & 0.7736 & 0.7840 & \\
\texttt{GPT4o\_mini} & 0.7684 & 0.7925 & 0.7711 & \\
\texttt{ESG-BERT} & 0.8360 & 0.6792 & 0.7050 & \color{black}\\
\hline
\end{tabular}
\vspace{2mm}
\caption{Performance of the fine-tuned models (original and synthetic data).}
\label{table:FT-synthetic}
\end{table}

In conclusion, our experiments show that LLMs, particularly open lightweight models that can run in low-resource settings, achieve high performance in classifying text related to ESG activities when fine-tuned on high-quality datasets. In contrast, more traditional encoder-based approaches such as \texttt{ESG-BERT} are less effective. These findings also provide further evidence supporting the effectiveness of synthetic data in specific domains. Notably, the most comprehensive version of the ESG-Activities benchmark training set, which combines both manually annotated and synthetic data, further improves classification performance. For example, \texttt{Llama2\_7B}, the best-performing model, achieved a remarkable 14.8\% increase in F1 score, rising from 70.1\% to 84.9\% with the inclusion of synthetic data.

\subsection{Training and Inference Time}
Training times varied significantly across models, reflecting differences in architecture, size, and the complexity of the fine-tuning processes.  
Table \ref{tab:inference_training_times} presents the training and inference times of the different models. We do not report the fine-tuning time for \texttt{GPT4o\_mini} because it was handled by the OpenAI service.

\begin{table}[h!]
\centering
\small
\begin{tabular}{lcccc}
\hline
\textbf{Model} & \textbf{Inference Time} & \textbf{Training Time} & \textbf{Training Time}\\ 
 & \textbf{(s)} & \textbf{Original (s)} & \textbf{Original+Synthetic} \\
 &&& \textbf{(s)}\\
\hline 
\texttt{Gemma\_2B} & 3.24 & 115 & 828 \\ 
\texttt{Gemma\_7B} & 2.18 & 120 & 976 \\ 
\texttt{RecGemma\_2B} & 2.51 & 130 & 1011\\
\texttt{Llama3\_8B} & 3.74 & 226 & 1096 \\ 
\texttt{Llama2\_7B} & 2.14 & 132 & 1046 \\ 
\texttt{Llama\_3B} & 3.61 & 181 & 1534  \\ 
\texttt{Mistral\_7B} & 10.89 & 703 & 5888 \\ 
\texttt{ESG-BERT}& 1.11 & 78 & 124 \color{black}\\ 
\texttt{GPT4o\_mini}& 4.46 & NA & NA \\ 
\hline
\end{tabular}
\caption{Training and inference time for the nine models.}
\label{tab:inference_training_times}
\end{table}

Notably, models from the Gemma family exhibit faster training times compared to the Llama models. Among the models analyzed, \texttt{Mistral\_7B} stands out with a significantly longer training time of about 96 minutes, nearly seven times that of \texttt{Gemma\_2B}. Regarding inference times, \texttt{Llama2\_7B} is the fastest model, followed by \texttt{Gemma\_7B} and \texttt{RecurrentGemma\_2B}.

\subsection{Cost Analysis}
The computational demands of training and inference processes can have a significant impact on the overall cost of a project. 

\begin{table}[h!]
\centering
\begin{tabular}{lccc}
\hline
\textbf{Model} & \textbf{original} & \textbf{orig+syntetic}\\ 
 & \textbf{(\$)} & \textbf{(\$)}\\ 
\hline 
\texttt{GPT4o\_mini}& 0.14 & 0.86 \\
\texttt{ESG-BERT} & 0.20 & 0.31 \\ 
\texttt{Gemma\_2B} & 0.29 & 2.07 \\ 
\texttt{Gemma\_7B} & 0.30 & 2.44 \\ 
\texttt{RecurrentGemma\_2B} & 0.33 & 2.52 \\
\texttt{Llama2\_7B} & 0.33 & 2.62 \\ 
\texttt{Llama3\_8B} & 0.56 & 2.74 \\ 
\texttt{Llama\_3B} & 0.45 & 3.84  \\ 
\texttt{Mistral\_7B} & 1.76 & 14.72 \\
\hline
\end{tabular}
\caption{Fine-tuning costs}
\label{tab:costs_table}
\end{table}

Table \ref{tab:costs_table} reports the costs associated with fine-tuning the nine models on both the original dataset and the full version, which includes synthetic data. All models, except \texttt{GPT4o\_mini}, were trained on an NVIDIA A100 GPU. The hourly rental rate for an A100 machine was approximately \$9 at the time of this study. 
Differently, the fine-tuning of \texttt{GPT4o\_mini} was conducted using OpenAI's standard API offering, which, at the time of writing, allowed fine-tuning at a cost of \$3.00 per 1 million tokens. 
For these reasons, although \texttt{GPT4o\_mini} is the largest model in terms of the number of parameters, it is the cheapest to train using OpenAI's services. However, this training cost must be combined with the fees required to deploy and maintain the model on OpenAI's infrastructure, which is currently \$1.62 an hour. 
 The open LLMs yielded similar costs, with the exception of \texttt{Mistral\_7B}, which was clearly more expensive. \texttt{ESG-BERT} was the most cost-effective due to its small size; however, its performance may not be sufficient to justify the savings.

\section{Related work}
\label{sec6:sota}

This section examines key advancements in the development and application of LLMs in the ESG domain.  
LLMs have demonstrated remarkable capabilities in processing textual data at scale, enabling the accurate extraction and synthesis of information from diverse sources, including research papers~\cite{bolanos2024artificial,cadeddu2024optimizing}, patents~\cite{kosonocky2024mining}, social media posts~\cite{yang2024mentallama}, medical records~\cite{omiye2024large},  news articles~\cite{motta2024capturing}, 
legal texts~\cite{savelka2023unreasonable}, knowledge graphs~\cite{tsaneva2025knowledge}, structured representation of scientific knowledge~\cite{meloni2025exploring}, and financial documents~\cite{wu2023bloomberggpt}. 
Recent LLM families, including GPT~\cite{openai2023gpt4o}, Llama~\cite{llama}, Mistral~\cite{mistral}, and Gemma~\cite{gemma}, have demonstrated outstanding capabilities in analyzing and classifying ESG and financial data. Their proficiency in processing vast amounts of unstructured information positions them as powerful tools for enhancing ESG reporting, regulatory compliance, and data-driven decision-making. 
However, the standard versions of these models, when used with simple prompting, do not always achieve optimal performance. Consequently, many researchers focus on developing domain-specific adaptations by fine-tuning open models on ESG and financial data. 
For instance, Yang et al.~\cite{yang2023fingpt} introduced FinGPT, an open-source LLM designed for the financial sector. Initially trained on a general-purpose corpus, FinGPT was then fine-tuned using domain-specific datasets, including financial news, earnings reports, and real-time market data. 
Despite the availability of several versions of FinGPT, none proved suitable for our purposes. In particular, we tested a multitask version of the model configured for output generation, but it turned out to be entirely inadequate for our ESG activity classification task, as it failed to produce any concrete or usable outputs. This limitation is largely due to the nature of the ESG classification task, which requires precise, binary or categorical decisions aligned with specific regulatory taxonomies, rather than the open-ended, natural language responses that FinGPT is optimized to generate. 
Another domain-specific solution is ESG-BERT~\cite{mukherjee2020esg}, a specialized variant of BERT that was discussed in Section~\ref{ref:models_list} and included in our experiments. This model achieved subpar performance on the task addressed in this paper, mainly because of its low recall.
Finally, FinBERT~\cite{huang2022finbert} is a domain-specific adaptation of BERT~\cite{kenton2019bert} designed for financial text analysis. It is pre-trained on a vast corpus of financial documents, including regulatory filings, corporate announcements, and financial articles, before undergoing fine-tuning for specific tasks.  
We attempted to adapt FinBERT for our binary classification task by removing its task-specific output layer and replacing it with a binary classification head, following the same approach applied to ESG-BERT. However, no publicly available checkpoint without a classification head could be found, and this adaptation resulted in very poor performance on ESG-Activities (F1 < 20\%). Based on these results, we decided not to include FinBERT in the final performance tables.
\color{black}
Several approaches have been proposed for ESG sentiment classification, which involves categorizing text as either positive/opportunity or negative/risk from an ESG perspective. Notably, the ML-ESG-2 shared task~\cite{kannan2023textual}, co-located with the FinNLP workshop at IJCNLP-AACL 2023, has played a key role in advancing methodologies in this field. To tackle this challenge, researchers have leveraged both encoder-only and decoder-only transformer models, including fine-tuned versions of FinBERT~\cite{mishra2023esg} and LLMs such as Llama 2 and Dolly~~\cite{zou2023esgreveal}. Similar to previous approaches, we leverage and evaluate a variety of fine-tuned language models. However, the task addressed in this paper is more fine-grained than merely detecting a positive or negative sentiment.
To tackle the challenges of long contexts and data recency in LLMs, researchers have started incorporating the RAG paradigm \cite{lewis2020retrieval}, which improves response accuracy by dynamically retrieving relevant information. For instance, ESGReveal \cite{zou2023esgreveal} is a recent method for extracting and analyzing ESG data from corporate reports that utilizes an LLM augmented with RAG. The system includes an ESG metadata module for precise querying, a preprocessing module for database construction, and an LLM agent for extracting information. Like our approach, ESGReveal leverages an RAG architecture. However, its primary focus is on extracting quantitative data from documents, such as Scope 1 and Scope 2 emissions. In contrast, our work is centred on identifying portions of documents that align with the activities described in the ESG taxonomy. Furthermore, ESGReveal utilizes zero-shot learning LLMs with basic prompting, whereas we fine-tune LLMs to optimize their performance for our specific task.
Efforts have also been made to develop a question-answering system in the ESG domain using a pipeline based on Knowledge Graph-based Retrieval Augmented Generation (KG-RAG) \cite{gupta2024knowledge}. Unlike traditional RAG, which retrieves information from textual sources, KG-RAG leverages knowledge graphs \cite{peng2023knowledge}, which are structured representations of information that capture relationships between entities in a graph format. Similarly, Angioni et al.~\cite{angioni2024exploring} employ a combination of linguistic pattern analysis and transformer models to extract a knowledge graph from a corpus of news articles, aiming to capture key trends related to ESG aspects. 
In this work, we focus on the more specific task of classifying text based on ESG activities. Consequently, we rely exclusively on action descriptions from the ESG taxonomy rather than employing a more complex knowledge representation.

\section{Conclusion}
\label{sec7:conclusion}

In this paper, we investigate the ability of current-generation LLMs to identify text related to ESG activities. Specifically, we assess strategies for optimizing these models using high-quality data and explore the feasibility of leveraging a combination of original and synthetically generated data. To support this study, we introduce ESG-Activities, a novel benchmark comprising 1,325 labelled text segments classified according to the EU ESG taxonomy.

Our results demonstrate that fine-tuned models significantly outperform zero-shot approaches, confirming that domain adaptation plays a crucial role in enhancing ESG text classification.
Moreover, the inclusion of synthetic data proved to be a valuable strategy, yielding substantial performance gains across multiple model architectures. This finding suggests that synthetic data augmentation can serve as an effective method to overcome data scarcity issues, particularly in specialized domains such as ESG analysis. Another key takeaway from our experiments is that relatively small, open-source models can achieve competitive performance levels when fine-tuned with high-quality, domain-specific data. This insight has significant implications for organizations looking to deploy cost-effective AI solutions for ESG analysis or those that need to run models locally due to privacy and data security concerns, which are particularly common in the financial domain.

This study opens several promising avenues for future research. First, we plan to explore alternative data generation techniques, including human-in-the-loop approaches, to further refine synthetic data for ESG applications. Second, while our work focused on ESG-related sentence classification, further research is needed to evaluate the models' performance on other NLP tasks, such as ESG sentiment analysis and risk assessment. 
Finally, future studies should examine how these models can be effectively deployed in real-world financial decision-making processes, ensuring that AI-driven ESG analysis aligns with regulatory requirements and industry standards.

\begin{acks}
This work is supported by the Italian Ministry of University and Research (MUR) within the PRIN2022-ISALDI: Interpretable Stock Analysis Leveraging Deep Multimodal Models (CUP: E53D23008150006).
\end{acks}

\bibliographystyle{ACM-Reference-Format}
\bibliography{biblio}









\end{document}